\documentclass[10pt,twocolumn]{article}

\usepackage{cvpr}
\usepackage{times}
\usepackage{epsfig}
\usepackage{graphicx}
\usepackage{amsmath}
\usepackage{adjustbox}
\usepackage{balance}

\usepackage{url}
\PassOptionsToPackage{hyphens}{url}
\usepackage[pagebackref=true,breaklinks=true,colorlinks,bookmarks=false]{hyperref}

\usepackage[ruled,vlined]{algorithm2e}

\cvprfinalcopy

\ifcvprfinal\fi
\begin{document}

\title{DeepFake Detection by Analyzing Convolutional Traces}

\author{Luca Guarnera \\
University of Catania - iCTLab\\
Catania, Italy\\
{\tt\small luca.guarnera@unict.it}
\and
Oliver Giudice\\
University of Catania \\
Catania, Italy\\
{\tt\small giudice@dmi.unict.it}
\and
Sebastiano Battiato \\
University of Catania - iCTLab\\
Catania, Italy\\
{\tt\small battiato@dmi.unict.it}
}

\maketitle
\thispagestyle{empty}

\begin{abstract}
The  Deepfake  phenomenon  has  become  very  popular nowadays  thanks  to  the  possibility to create incredibly realistic images using deep learning tools, based mainly on ad-hoc Generative Adversarial Networks (GAN). In this work we focus on the analysis of Deepfakes of human faces with the objective of creating a new detection method able to detect a forensics trace hidden in images: a sort of fingerprint left in the image generation process. The proposed technique, by means of an Expectation Maximization (EM) algorithm, extracts a set of local features specifically addressed to model the underlying convolutional generative process. Ad-hoc validation has been employed through experimental tests with naive classifiers on five different architectures (GDWCT,  STARGAN, ATTGAN, STYLEGAN, STYLEGAN2) against the CELEBA dataset as ground-truth for non-fakes. Results demonstrated the effectiveness of the technique in distinguishing the different architectures and the corresponding generation process.
\end{abstract}

\section{Introduction}
\label{sec:intro}

\begin{figure*}[t]
    \centering
    \frame{\includegraphics[width=17cm]{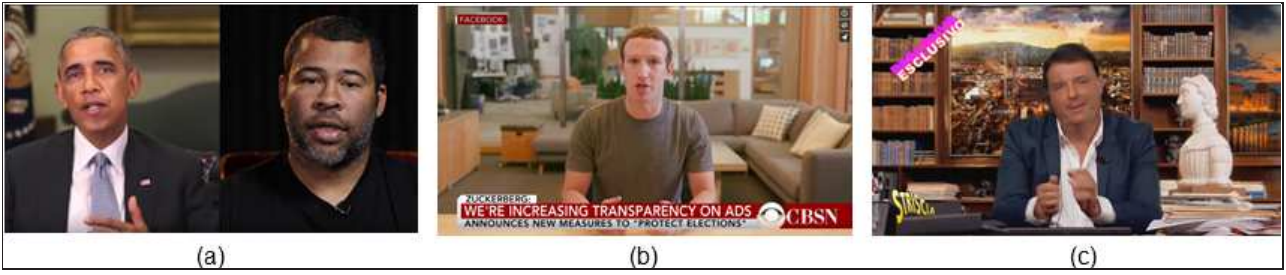}}
    \caption{Examples of Deepfake: (a) Obama (Buzzfeed in collaboration with Monkeypaw Studios); (b) Mark Zuckerberg (Bill Posters and Daniel Howe in partnership with advertising company Canny); (c) Matteo Renzi (the italian TV program ``Striscia la Notizia").}
    \label{fig:1}
\end{figure*}

\begin{figure}[t!]
    \centering
    \includegraphics[width=8cm]{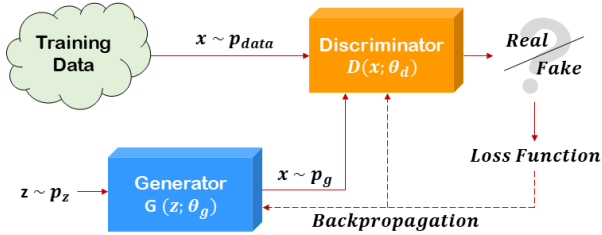}
    \caption{Simplified description of a GAN learning framework.}
    \label{fig:genericalGAN}
\end{figure}

One of the phenomena that is rapidly growing is the well-known Deepfake: the possibility to automatically generate and/or alter/swap a person's face in images and videos using algorithms based on \textit {Deep Learning} technology. It is possible to generate excellent results by creating new multimedia contents that cannot be easily recognized as real or fake by human eye. Then, the term Deepfake refers to all those multimedia contents synthetically altered or created by means of machine learning generative models. 

Various examples of Deepfake, involving celebrities, are easily discoverable on the web: the insertion of Nicholas Cage~\footnote{\href{https://www.youtube.com/watch?v=-yQxsIWO2ic}{https://www.youtube.com/watch?v=-yQxsIWO2ic}} in movies where he did not act like ``Fight Club" and ``The Matrix" or the impressive video in which Jim Carrey~\footnote{\href{https://www.youtube.com/watch?v=Dx59bskG8dc}{https://www.youtube.com/watch?v=Dx59bskG8dc}} plays Shining in place of Jack Nicholson. Other more worrying examples are the video of ex US President Barack Obama (Figure~\ref{fig:1}(a)), created by Buzzfeed~\footnote{\href{https://www.youtube.com/watch?v=cQ54GDm1eL0}{https://www.youtube.com/watch?v=cQ54GDm1eL0}} in collaboration with Monkeypaw Studios, or the video in which Mark Zuckerberg~\footnote{\href{https://www.youtube.com/watch?v=NbedWhzx1rs}{https://www.youtube.com/watch?v=NbedWhzx1rs}} (Figure~\ref{fig:1}(b)) claims a series of statements about his platform ability to steal users' data. 
Even in Italy, in September 2019, the satirical TV program ``Striscia La Notizia"~\footnote{\href{https://www.striscialanotizia.mediaset.it/video/il-fuorionda-di-matteo-renzi_59895.shtml}{https://www.striscialanotizia.mediaset.it/Matteo-Renzi}} showed a video of the ex-premier Matteo Renzi talking about his colleagues in a ``not so respectful" way (Figure~\ref{fig:1} (c)). Indeed, Deepfakes may have serious repercussions on the authenticity of the news spread by the mass-media while representing  a new threat for politics, companies and individual privacy. In this dangerous scenario, tools are needed to unmask the Deepfakes or just detect them.

Several big companies have decided to take action against this phenomenon: Google has created a database of fake videos~\cite{rossler2019faceforensics++} to support researchers who are developing new techniques to detect them, while Facebook and Microsoft have launched the Deepfake Detection Challenge initiative~\footnote{\href{https://deepfakedetectionchallenge.ai/}{https://deepfakedetectionchallenge.ai/}}.

In this paper a new Deepfake detection method will be introduced focused on images representing human faces. At first an Expectation Maximization (EM) algorithm~\cite{moon1996expectation}, extracts a set of local features specifically addressed to model the convolutional traces that could be found in images. Then, naive classifiers were trained to discriminate between authentic images and images generated by the five most realistic architectures as today (GDWCT,  STARGAN, ATTGAN, STYLEGAN, STYLEGAN2). Experimental results demonstrated that the information modelled by EM is related to the specific architecture that generated the image thus giving the overall detection solution explainability, being also of great value for forensic investigations (e.g., camera model identification techniques of image forensics). Moreover, a multitude of experiments will be presented not only to demonstrated the effectiveness of the technique but also to demonstrate to be un-comparable with state-of-the-art: tests were carried out on an almost-in-the-wild dataset with images generated by five different techniques with different image sizes. As today, all proposed technique work with specific image sizes and against at most one GAN technique.

The remainder of this paper is organized as follows: Section~\ref{sec:related} presents some Deepfake generation and detection methods. The proposed detection technique is explained in Section \ref{sec:approach} as regards the feature extraction phase while the classification phase and experimental results are reported in Section \ref{sec:results}. Finally, Section~\ref{sec:conclusion} concludes the paper with insights for future works.

\section{Related Works}
\label{sec:related}

Deepfakes are generally created by techniques based on Generative Adversarial Networks (GANs) firstly introduced by Goodfellow et al.~\cite{goodfellow2014generative}. Authors proposed a new framework for estimating generative models via an adversarial mode in which two models simultaneously train: a generative model $G$, that captures the data distribution, and a discriminative model $D$, able to estimate the probability that a sample comes from the training data rather than from $G$. The training procedure for $G$ is to maximize the probability of $D$ making a mistake thus resulting to a min-max two-player game.
Mathematically, the generator accepts a random input $z$ with density $p_z$ and returns an output $x = G (z, \Theta_g)$ according to a certain probability distribution $p_g$ ($\Theta_g$ represent the parameters of the generative model). The discriminator, $D(x, \Theta_d)$ computes the probability that $x$ comes from the distribution of training data $p_{data}$ ($\Theta_d$ represents the parameters of the discriminative model). The overall objective is to obtain a generator, after the training phase, which is a good estimator of $p_{data}$. When this happens, the discriminator is ``deceived" and will no longer be able to distinguish the samples from $p_{data}$ and $p_g$; therefore $p_g$ will follow the targeted probability distribution, i.e. $p_{data}$. Figure~\ref{fig:genericalGAN} shows a simplified description of a GAN framework.
In the case of Deepfakes, $G$ can be thought as a team of counterfeiters trying to produce fake currency, while $D$ stands to the police, trying to detect the malicious activity. $G$ and $D$ can be implemented as any kind of generative model, in particular when deep neural networks are employed results become extremely accurate.
Through recent years, many GAN architectures were proposed for different applications e.g., image to image translation~\cite{zhu2017unpaired}, image super resolution~\cite{ledig2017photo}, image completion~\cite{iizuka2017globally}, and text-to-image generation~\cite{reed2016generative}. 

\subsection{Deepfake Generation Techniques}

An overview on Media forensics with particular focus on Deepfakes has been recently proposed in~\cite{verdoliva2020media}.

STARGAN is a method capable of performing image-to-image translations on multiple domains using a single model. Proposed by Choi et al.~\cite{choi2018stargan} was trained on two different types of face datasets: CELEBA~\cite{liu2015deep} containing 40 labels related to facial attributes such as hair color, gender and age, and RaFD dataset~\cite{langner2010presentation} containing 8 labels corresponding to different types of facial expressions (``happy", ``sad", etc.). Given a random label as input, such as hair color, facial expression, etc., STARGAN is able to perform an image-to-image translation operation. Results have been compared with other existing methods~\cite{li2016deep, perarnau2016invertible, zhu2017unpaired} and showed how STARGAN manages to generate images of superior visual quality.

Style Generative Adversarial Network, namely STYLEGAN~\cite{karras2019style}, changed the generator model of STARGAN by means of mapping points in latent space to an intermediate latent space which controls the \textit{style} output at each point of the generation process. Moreover the introduction of noise as a source of variation in those mentioned points demonstrates to achieve better results. Thus, STYLEGAN is capable not only of generating impressively photorealistic and high-quality photos of faces, but also offers control parameters in terms of the overall \textit{style} the generated image at different levels of detail. While being able to create realistic pseudo-portraits, small details might reveal the fakeness of generated images.  To correct those imperfections in STYLEGAN, Karras et al. made some improvements to the generator (including re-designed normalization, multi-resolution, and regularization methods) proposing STYLEGAN2~\cite{karras2019analyzing}. 

Instead of imposing constraints on latent representation, He et al.~\cite{he2019attgan}, proposed a new technique called ATTGAN in which an attribute classification constraint is applied to the generated image, in order to guarantee only the correct modifications of the desired attributes. The authors used CELEBA~\cite{liu2015deep} and LFW~\cite{huang2008labeled} datasets, and performed various tests comparing ATTGAN with VAE/GAN~\cite{larsen2015autoencoding}, IcGAN~\cite{perarnau2016invertible} and STARGAN~\cite{choi2018stargan}, Fader Networks~\cite{lample2017fader}, Shen et al.~\cite{shen2017learning} and CycleGAN~\cite{zhu2017unpaired}. Achieved results showed that ATTGAN exceeds the state of the art on the realistic modification of facial attributes. 

The latter style transfer approach worth to be mentioned is the work of Cho et al.~\cite{cho2019image}, where they propose a group-wise deep whitening-and coloring method (GDWCT) for a better styling capacity. They used CELEBA~\cite{liu2015deep}, Artworks~\cite{zhu2017unpaired}, cat2dog~\cite{lee2018diverse}, Ink pen and watercolor classes from Behance Artistic Media (BAM)~\cite{wilber2017bam}, and Yosemite datasets~\cite{zhu2017unpaired} as dataset. GDWCT has been compared with various cutting-edge methods in image translation and style transfer improving not only computational efficiency but also quality of generated images.

In this paper, the five most famous and effective architectures in state-of-the-art for face Deepfakes were taken into account: STARGAN~\cite{choi2018stargan}, STYLEGAN~\cite{karras2019style}, STYLEGAN2~\cite{karras2019analyzing}, ATTGAN~\cite{he2019attgan} and GDWCT~\cite{cho2019image}. As described above, they are different in goals and structure. Table~\ref{tab:dataAndNetwork_} resumes the differences of the techniques in terms of image size, dataset and type of input, goal and architecture structure.

\subsection{Deepfake detection methods}

Being able to understand if an image is the result of a generative Neural Network process turns out to be a complicated problem, even for human eyes. However, the problem of authenticating an image (or specifically a digital image) is not new \cite{battiato2016multimedia,piva2013,stemm2013}. Many works try to reconstruct the history of an image\cite{Giudice2017}; others try to identify the anomalies, such as the study on the analysis of interpolation effects through CFA (Color Filtering Array)~\cite{popescu2005exposing}, analyzing compression parameters~\cite{battiato2009digital, galvan2014first, giudice20191}, etc. Given the peculiarity of Deepfakes, state-of-the-art image analysis methods tend to fail and more refined ones are needed.

Thanks to a new discriminator that uses ``contrastive loss" it is possible to find the typical characteristics of the synthesized images generated by different GANs and therefore detect such fake images by means of a classifier.
Rossler et al.~\cite{rossler2019faceforensics++} proposed an automated benchmark for fake detection, based mainly on four manipulation methods: two computer graphics-based methods (Face2Face~\cite{thies2016face2face}, FaceSwap~\footnote{\href{https://github.com/MarekKowalski/FaceSwap/}{https://github.com/MarekKowalski/FaceSwap/}}) and 2 learning-based approaches (DeepFakes~\footnote{\href{https://github.com/deepfakes/faceswap/}{https://github.com/deepfakes/faceswap/}}, NeuralTextures~\cite{thies2019deferred}). They addressed the problem of fake detection as a binary classification problem for each frame of manipulated videos, considering different techniques present in the state of the art \cite{afchar2018mesonet, bayar2016deep, chollet2017xception, cozzolino2017recasting, fridrich2012rich, rahmouni2017distinguishing}. 

Zhang et al.~\cite{zhang2019detecting} proposed a method to classify Deepfakes considering the spectra of the frequency domain as input. The authors proposed a GAN simulation framework, called AutoGAN, in order to emulate the process commonly shared by popular GAN models. Results obtained by the authors achieved very good performances in terms of binary classification between authentic and fake images. Also Durall et al.~\cite{durall2019unmasking} presented a method for Deepfakes detection based on the analysis in the frequency domain. The authors combined high-resolution authentic face images from different public datasets (CELEBA-HQ data set~\cite{karras2017progressive}, Flickr-Faces-HQ data set~\cite{karras2019style}) with fakes ($100$K Faces project~\footnote{\href{https://generated.photos/}{https://generated.photos/}}, this person does not exist~\footnote{\href{https://thispersondoesnotexist.com/}{https://thispersondoesnotexist.com/}}), creating a new dataset called Faces-HQ. By means of naive classifiers they obtained good results in terms of overall accuracy.

Differently from described approaches, in this paper the possibility to capture the underlying traces of a possible Deepfake is investigated by employing a sort of reverse engineering of the last computational layer of a given GAN architecture. This method will give explainability to the predictions of Deepfakes being of great value for forensic investigations: not only it is able to classify an image as fake but also can predict the most probable technique used for generation being in this way similar to camera model detection in image forensics analysis \cite{battiato2016multimedia}. The underlying idea of the technique is to find the main periodic components (e.g. transpose computational layer) on generated images. A similar strategy was proposed some time ago in a seminal paper of Popescu et al.~\cite{popescu2005exposing} devoted to point out the presence of digital forgeries in CFA interpolated images. Another difference from state-of-the-art is the working scenario: the proposed technique demonstrates to achieve good results in a almost-in-the-wild scenario with images generated by five different techniques and image sizes.

\begin{table*}[t]
\centering
\begin{adjustbox}{max width=\textwidth}
\begin{tabular}{|c|c|c|c|c|c|}
\hline
\textbf{Method}    & \textbf{\begin{tabular}[c]{@{}c@{}}Number of \\ images \\ generated\end{tabular}} & \textbf{Size} & \textbf{\begin{tabular}[c]{@{}c@{}}Data input \\ to the network\end{tabular}} & \textbf{Goal of the network}                                                                                                                      & \textbf{\begin{tabular}[c]{@{}c@{}}Kernel size of the \\ latest Convolution \\ Layer\end{tabular}} \\ \hline
\textbf{GDWCT~\cite{cho2019image}}    & 3369                                                                              & 216x216       & CELEBA                                                                        & \begin{tabular}[c]{@{}c@{}}Improves the styling \\ capability\end{tabular}                                                                    & 4x4                                                                                                  \\ \hline
\textbf{STARGAN~\cite{choi2018stargan}}   & 5648                                                                              & 256x256       & CELEBA                                                                        & \begin{tabular}[c]{@{}c@{}}Image-to-image translations\\ on multiple domains \\ using a single model\end{tabular}                                 & 7x7                                                                                   \\ \hline
\textbf{ATTGAN~\cite{he2019attgan}}    & 6005                                                                              & 256x256       & CELEBA                                                                        & \begin{tabular}[c]{@{}c@{}}Transfer of face attributes \\ with classification constraints\end{tabular}                                            & 4x4                                                                                  \\ \hline
\textbf{STYLEGAN~\cite{karras2019style}}  & 9999                                                                              & 1024x1024     & \begin{tabular}[c]{@{}c@{}}CELEBA-HQ\\ FFHQ\end{tabular}                      & \begin{tabular}[c]{@{}c@{}}Transfer semantic content from a \\ source domain to a target domain\\ characterized by a different style\end{tabular} & 3x3                                                                                           \\ \hline
\textbf{STYLEGAN2~\cite{karras2019analyzing}} & 3000                                                                              & 1024x1024     & FFHQ                                                                          & \begin{tabular}[c]{@{}c@{}}Transfer semantic content from a \\ source domain to a target domain\\ characterized by a different style\end{tabular} & 3x3                                                                                        \\ \hline

\end{tabular}
\end{adjustbox}
\caption{Details of Deepfake GAN architectures employed for analysis. For each one is reported: all images generated, the generated image sizes, the original input used to train the neural network, the goal of the network and the kernel size of last convolutional layer.}
\label{tab:dataAndNetwork_}
\end{table*}

\section{Extracting Convolutional Traces}
\label{sec:approach}

The most common and effective technical solutions able to generate Deepfakes are the Generative Adversarial Networks specifically deep ones. For all the techniques described before, the generator $G$ is composed of Transpose Convolution layers~\cite{radford2015unsupervised}. In Neural Networks like CNNs, Convolution operations apply a filter, namely kernel, to the input multidimensional array. After each convolution layer a pooling operation is needed to reduce output dimensional size w.r.t. input. On the other hand, in generative models the Transpose Convolution Layers are employed. They also apply kernels to input but they act inversely in order to obtain an output larger but proportional to the input dimensions.

The starting idea of the proposed approach is that local correlation of pixels in Deepfakes are  dependent exclusively on the operations performed by all the layers present in the GAN which generate it; specifically the (latter) transpose convolution layers. In order to find these trace, unsupervised machine learning techniques were taken into account. Indeed, different unsupervised learning techniques aim at creating clusters containing instances of the input dataset with high similarity between instances of the same cluster while having high dissimilarity between instances belonging to different clusters. These clusters can represent the ``hidden" structure of the dataset analyzed. Therefore, the clustering technique must estimate which are the parameters of the distributions that most likely generated the training samples. Based on this principle, an Expectation Maximization (EM) algorithm~\cite{moon1996expectation} was employed in order to define a conceptual mathematical model able to capture the pixel correlation of the images (e.g. spatially). The result of EM is a feature vector representing the structure of the Transpose Convolution Layers employed during the generation of the image, encoding in some sense is such images if a Deepfake or not.

The initial goal is to extract a description, from input image $I$, able to numerically represent the local correlations between each pixel in a neighbourhood. This can be done by means of convolution with a kernel $k$ of $N \times N$ size: 

\begin{equation}
	\label{eq:eqconv}
	I[x,y] = \sum\limits_{s,t=-\alpha}^\alpha k_{s,t}*I[x+s,y+t] 
\end{equation}

In Equation~\ref{eq:eqconv}, the value of the pixel $I[x,y]$ is computed considering a neighborhood of size $N \times N$ of the input data. It is clear that the new estimated information $I[x,y]$ mainly depends on the kernel used in the convolution operation, which establishes a mathematical relationship between the pixels. For this reason, our goal is to define a vector $k$ of size $N \times N$ able to capture this hidden and implicit relationship which characterize of forensic trace we want to exploit. 

Let's assume that the element $I[x,y]$ belongs to one of the following models:

\begin{itemize}
    \item $M_{1}$: when the element $I[x,y]$ satisfies Equation~\ref{eq:eqconv};
    \item $M_{2}$: otherwise.
\end{itemize}

The EM algorithm is employed with its two different steps:

\begin{enumerate}
    \item \textbf{Expectation step}: computes the (density of) probability that each element belongs to model ($M_{1}$ or $M_{2}$);
    \item \textbf{Maximization step}: estimates the (weighted) parameters based on the probabilities of belonging to instances of ($M_{1}$ or $M_{2}$).
\end{enumerate}

Let's suppose that $M_{1}$ and $M_{2}$ have different probability distributions with $M_{1}$  Gaussian distribution with zero mean and unknown variance  and $M_{2}$ uniform. In the Expectation step, the Bayes rule that $I[x, y]$ belongs to the model $M_{1}$ is computed as follows:

\begin{equation}
	\label{eq:Bayes}
	\begin{split}
	    Pr\{I[x,y] \in M_{1} \mid I[x,y]\} = \\
		= \frac{Pr\{I[x,y] \mid I[x,y] \in M_{1}\}*Pr\{I[x,y] \in M_{1}\}}{\sum\limits_{i=1}^2{Pr\{I[x,y] \mid I[x,y] \in M_{i}\}*Pr\{I[x,y] \in M_{i}\}}}
	\end{split}
\end{equation}

where the probability distribution of $M_{1}$ which represents the probability of observing a sample $I[x,y]$, knowing that it was generated by the model $M_{1}$ is:

\begin{equation}
	\label{eq:Bayes2}
	\begin{split}
	Pr\{I[x,y] \mid I[x,y] \in M_{1}\} =
	\frac{1}{\sigma\sqrt{2\pi}}e^{-\frac{(R[x,y])^2}{2\sigma^2}}
	\end{split}
\end{equation}
where
\begin{equation}
	\label{eq:BayesX}
	\begin{split}
	R[x,y]=\bigg|I[x,y]-\sum\limits_{s,t=-\alpha}^\alpha{k_{s,t}I[x+s,y+t]}\bigg|
	\end{split}
\end{equation}.

The variance value $\sigma^2$, which is still unknown, is then estimated in the Maximization step. Once defined if $I[x,y]$ belongs to model $M_{1}$ (or $M_{2}$), the values of the vector $\vec k$ are estimated using least squares method, minimizing the following:

\begin{equation}
	\label{eq:minimiz}
	E(\vec k) = \sum\limits_{x,y}w[x,y]\Bigg(I[x,y]-\sum\limits_{s,t=-\alpha}^\alpha{k_{s,t}I[x+s,y+t]} \Bigg)^2
\end{equation}

where $w \equiv Pr\{I[x,y] \in M_{1} \mid I[x,y]\}$ (\ref{eq:Bayes}). This error function (\ref{eq:minimiz}) can be minimized by computing the gradient of vector $\vec k$. The update of $k_{i, j}$ is carried out by computing the partial derivative of~(\ref{eq:minimiz}) as follows:

\begin{equation}
	\label{eq:derivata0}
	\frac{\partial E}{\partial k_{i,j}} = 0
\end{equation}

Hence, the following linear equations system is obtained:

\begin{equation}
	\label{eq:derivata}
	\begin{split}
	\sum\limits_{s,t=-\alpha}^\alpha k_{s,t}\Bigg(\sum\limits_{x,y}w[x,y]I[x+i, y+j]I[x+s, y+t]\Bigg) = \\
	=\sum\limits_{x,y}w[x,y]I[x+i, y+j]I[x, y]
	\end{split}
\end{equation}

The two steps of the EM algorithm are iteratively repeated. A pseudo-code description is provided in \textit{Algorithm 1:Expectation-Maximization}. The algorithm is applied to each channel of the input image (RGB color space). 

The obtained feature vector $\vec k$, has dimensions dependent to parameter $\alpha$. Note that the element $k_{0,0}$ will always be set equal to 0 ($k_{0,0}=0$). Thus, for example, if a kernel $k$ with $3 \times 3$ size is employed, the resulting $\vec k$ will be a vector of $24$ elements (since the values $k_{0,0}$ are excluded). This is obtained by concatenating the features extracted from each of the three RGB channels.

The computational complexity of the EM algorithm can be estimated to be linear in $d$ (the number of characteristics of the input data taken into consideration), $n$ (the number of objects) and $t$ (the number of iterations).

\begin{algorithm}[t]
	\caption{Expectation-Maximization Algorithm}
	\label{alg:Expectation_Maximization}
	\KwData{Image $I$}
	\KwResult{$\vec k$}
	Initialize $N$ //Kernel size \\
	Initialize $\sigma_0$ \\
	Set $\vec k$ random of size NxN \\
	Set $R, P, W$ matrices with $0$ values of the same size as $I$ \\
	Set $p_0$ as 1/size of the range of values of $I$ \\
	\For{$n = 1; \enspace n < 100 \enspace n+=1$}{ 
	    //Expectation Step \\
	    \For{$\forall$ values in $I$}{ 
	        $R[x,y]=\bigg|I[x,y]-\sum\limits_{s,t=-\alpha}^\alpha{k_{s,t}I[x+s,y+t]}\bigg|$ \\
	        $P[x,y] = \frac{1}{\sigma_n\sqrt{2\pi}}e^{-\frac{R[x,y]}{2\sigma_n^2}}$ \\
	        $W[x,y] = \frac{P[x,y]}{P[x,y]+p_0}$ \\
	    }
	    //Maximization Step \\
	    Calculate $k_{s,t}^{(n+1)}$ as shown in the formula~\ref{eq:derivata} \\
	}
\end{algorithm}

\section{Classification Phase and Results}
\label{sec:results}

\begin{figure*}[t!]
\begin{center}
\fbox{\rule{0pt}{2in} \rule{.0\linewidth}{0pt}\includegraphics[width=0.9\linewidth]{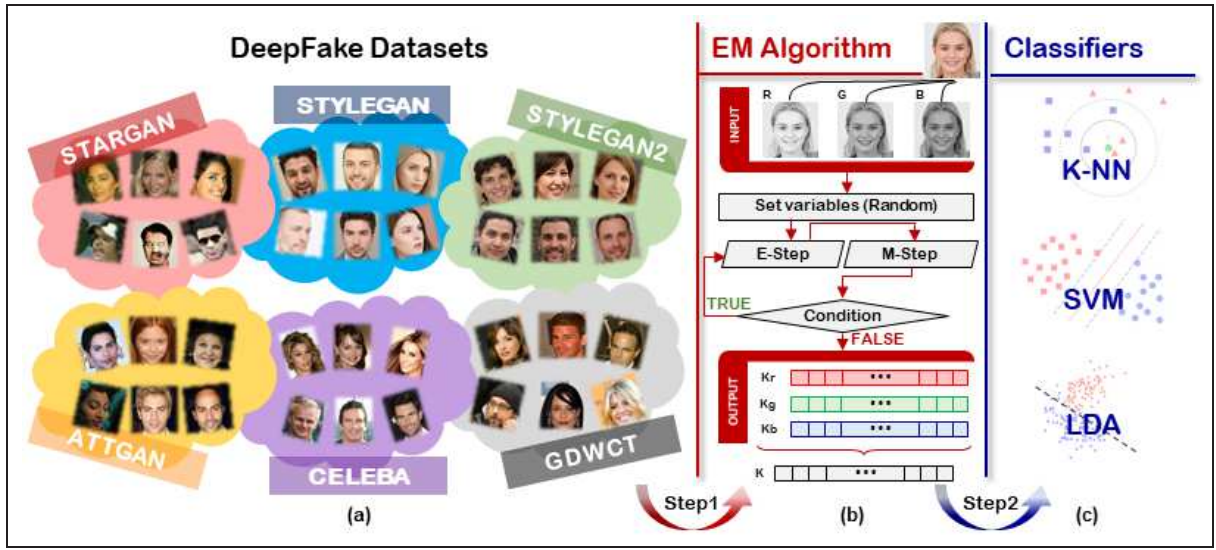}}
\end{center}
   \caption{Overall pipeline. (a) Datasets of real (CELEBA) and Deepfake images, (b) For each images in (a) features are extracted by means of EM algorithm; (c) types of classifiers used (K-NN, SVM, LDA).}
\label{fig:Pipeline}
\end{figure*}

Six datasets of images were taken into account for training and testing purposes: one containing only authentic face images of celebrities (CELEBA), and the others containing DeepFakes generated by five different GANs (STARGAN, STYLEGAN, STYLEGAN2, GDWCT, ATTGAN). For STYLEGAN and STYLEGAN2, images were downloaded from STYLEGAN~\footnote{\href{https://drive.google.com/drive/folders/1uka3a1noXHAydRPRbknqwKVGODvnmUBX}{https://drive.google.com/StyleGAN}} and STYLEGAN2~\footnote{\href{https://drive.google.com/drive/folders/1QHc-yF5C3DChRwSdZKcx1w6K8JvSxQi7}{https://drive.google.com/StyleGAN2}} respectively; while STARGAN, ATTGAN and GDWCT were employed in inference mode to generate their respective image datasets. An overview of the DeepFake data generated for each GAN is reported in the Table~\ref{tab:dataAndNetwork_}.

\begin{figure*}[t]
\begin{center}
\fbox{\rule{0pt}{2in} \rule{.0\linewidth}{0pt}\includegraphics[width=0.70\linewidth]{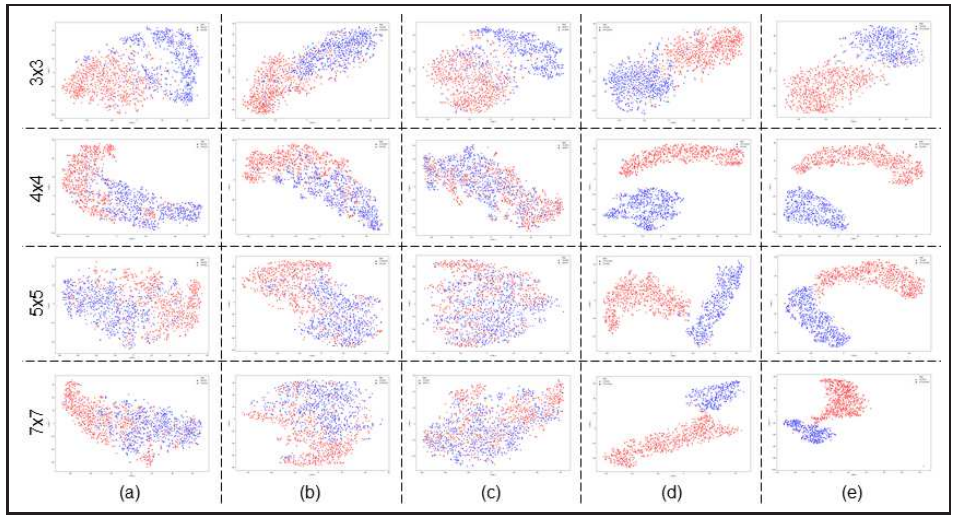}}
\end{center}
   \caption{Two-dimensional t-SNE representation (CELEBA: red; DeepNetwork: blue) of all kernel sizes for each classification task: (a) CELEBA – ATTGAN; (b) CELEBA – STARGAN; (c) CELEBA – GDWCT; (d) CELEBA – STYLEGAN; (e) CELEBA – STYLEGAN2.}
\label{fig:tsneAll}
\end{figure*}

The EM algorithm, as described in previous Section, was employed on the 6 datasets described above, in order to extract a feature vector $\vec k$ able to describe the convolutional traces left in images. EM was employed with kernels of increasing sizes ($ 3, 4, 5$ and $7$)~\footnote{Typical kernel size used by the latest Transpose Convolution Layers (which have a fundamental role in the creation of the Deepfake images) of the different GAN architectures}. The obtained feature vector was employed as input of different naive classifiers (K-NN, SVM and LDA) with different tasks: (i) discriminating authentic image from one specific GAN and (ii) discriminating authentic images from Deepfakes. The overall classification pipeline of the proposed approach is briefly summarized in Figure~\ref{fig:Pipeline}. 
Let's first analyse the discriminative power of the extracted feature vector in order to distinguish authentic images (CELEBA) from each of the considered GANs (CELEBA Vs STARGAN, CELEBA Vs STYLEGAN, CELEBA Vs STYLEGAN2, CELEBA Vs ATTGAN, CELEBA Vs GDWCT). Figure~\ref{fig:tsneAll} shows a visible representation by means of t-SNE~\cite{maaten2008visualizing}: in which it is possible to notice, how some categories of networks that create Deepfake can be ``linearly" separable from authentic samples. However in most case the separation is utterly clear.

Classification tests were carried out on the obtained feature vectors with, as expected from what seen from t-SNE representation, excellent results. All the classification results are reported in Table~\ref{tab:accuracyCELEBAvsSINGLEDEEPNETWORK}. In particular, it is possible to note that:

\begin{itemize}
    \item \textbf{CELEBA Vs ATTGAN} the maximum classification accuracy of $\textbf{92.67\%}$, was obtained with KNN - K = 3, and kernel size of $3x3$.
    \item \textbf{CELEBA Vs GDWCT}: the maximum classification accuracy of $\textbf{88.40\%}$, was obtained with KNN - K = 3,5,7, and kernel size of $3x3$.
    \item \textbf{CELEBA Vs STARGAN}: the maximum classification accuracy of $\textbf{93.17\%}$, was obtained with linear SVM, and kernel size of $7x7$.
    \item \textbf{CELEBA Vs STYLEGAN}: the maximum classification accuracy of $\textbf{99.65\%}$, was obtained with KNN - K = 3,5,7,9, and kernel size of $4x4$.
    \item \textbf{CELEBA Vs STYLEGAN2}: the maximum classification accuracy of $\textbf{99.81\%}$, was obtained with linear SVM, and kernel size of $4x4$.
\end{itemize}

The kernel size used by convolution layers in the neural networks represents one of the  elements to identify the forensic trace that we are looking for. Table~\ref{tab:dataAndNetwork_} shows the kernel size (and other information) of the neural networks that we have taken into account for our experiments.

\begin{table*}[t!]
\centering
  \begin{adjustbox}{max width=\textwidth}
\begin{tabular}{c|cccc|cccc|cccc|cccc|cccc|}
\cline{2-21}
                                     & \multicolumn{4}{c|}{\textbf{CELEBA Vs ATTGAN}}                 & \multicolumn{4}{c|}{\textbf{CELEBA Vs GDWCT}}                & \multicolumn{4}{c|}{\textbf{CELEBA Vs STARGAN}}                  & \multicolumn{4}{c|}{\textbf{CELEBA Vs STYLEGAN}}               & \multicolumn{4}{c|}{\textbf{CELEBA Vs STYLEGAN2}}            \\ \cline{2-21} 
\textbf{}                            & \multicolumn{4}{c|}{\textbf{Kernel Size}}                     & \multicolumn{4}{c|}{\textbf{Kernel Size}}                   & \multicolumn{4}{c|}{\textbf{Kernel Size}}                       & \multicolumn{4}{c|}{\textbf{Kernel Size}}                     & \multicolumn{4}{c|}{\textbf{Kernel Size}}                   \\
\textbf{}                            & \textbf{3x3}   & \textbf{4x4} & \textbf{5x5}   & \textbf{7x7} & \textbf{3x3}   & \textbf{4x4} & \textbf{5x5} & \textbf{7x7} & \textbf{3x3}   & \textbf{4x4}   & \textbf{5x5} & \textbf{7x7}   & \textbf{3x3} & \textbf{4x4}   & \textbf{5x5}   & \textbf{7x7} & \textbf{3x3} & \textbf{4x4}   & \textbf{5x5} & \textbf{7x7} \\ \hline
\multicolumn{1}{|c|}{\textbf{3-NN}}  & \textbf{92.67} & 86.50        & 84.50          & 85.33        & \textbf{88.40} & 73.17        & 73.00        & 74.33        & \textbf{90.50} & 89.00          & 88.67        & 85.17          & 93.00        & \textbf{99.65} & 98.26          & 99.55        & 96.99        & \textbf{99.61} & 98.75        & 97.77        \\
\multicolumn{1}{|c|}{\textbf{5-NN}}  & \textbf{92.00} & 86.50        & 84.83          & 86.17        & \textbf{88.40} & 75.67        & 74.17        & 76.67        & \textbf{88.83} & \textbf{88.83} & 88.17        & 85.00          & 93.00        & \textbf{99.65} & 98.26          & 99.32        & 97.39        & \textbf{99.61} & 98.21        & 97.55        \\
\multicolumn{1}{|c|}{\textbf{7-NN}}  & \textbf{91.00} & 87.67        & 85.33          & 85.67        & \textbf{88.40} & 76.67        & 71.33        & 78.67        & \textbf{89.33} & 89.17          & 88.00        & 84.83          & 93.50        & \textbf{99.65} & 98.07          & 99.09        & 97.39        & \textbf{99.42} & 98.21        & 97.55        \\
\multicolumn{1}{|c|}{\textbf{9-NN}}  & \textbf{90.83} & 87.67        & 84.83          & 86.50        & \textbf{87.70} & 76.83        & 71.17        & 79.00        & \textbf{89.33} & 89.17          & 87.50        & 84.67          & 92.83        & \textbf{99.65} & 98.07          & 99.32        & 97.19        & \textbf{99.42} & 98.39        & 97.10        \\
\multicolumn{1}{|c|}{\textbf{11-NN}} & \textbf{91.00} & 86.83        & 85.33          & 85.83        & \textbf{88.05} & 76.67        & 72.83        & 77.00        & \textbf{89.17} & 88.67          & 86.67        & 83.50          & 93.17        & \textbf{99.48} & 98.07          & 99.32        & 96.99        & \textbf{99.42} & 97.85        & 97.10        \\
\multicolumn{1}{|c|}{\textbf{13-NN}} & \textbf{91.00} & 87.17        & 84.50          & 85.33        & \textbf{87.87} & 75.33        & 73.50        & 77.17        & 88.33          & \textbf{89.33} & 87.50        & 83.50          & 93.50        & \textbf{99.48} & 98.07          & 99.09        & 97.39        & \textbf{99.22} & 97.67        & 97.10        \\
\multicolumn{1}{|c|}{\textbf{SVM}}   & \textbf{90.50} & 89.67        & 90.33          & 87.00        & \textbf{87.35} & 76.50        & 79.00        & 80.50        & 90.00          & 88.50          & 88.83        & \textbf{93.17} & 92.00        & 98.96          & \textbf{99.42} & 98.41        & 96.99        & \textbf{99.81} & 99.46        & 97.77        \\
\multicolumn{1}{|c|}{\textbf{LDA}}   & \textbf{89.50} & 88.50        & \textbf{89.50} & 87.17        & \textbf{87.52} & 76.00        & 79.33        & 81.67        & 89.67          & 87.83          & 88.83        & \textbf{90.00} & 92.50        & \textbf{99.31} & 98.84          & 99.09        & 96.79        & \textbf{99.61} & 99.10        & 97.77        \\ \hline
\end{tabular}
\end{adjustbox}
\caption{Overall accuracy between CELEBA vs. each one of the considered GANs. Results are presented w.r.t. all the different kernel sizes ($3x3$, $4x4$, $5x5$, $7x7$) and with different classifiers: KNN, with $k \in \{3,5,7,9,11,13\}$; Linear SVM, Linear Discriminant Analysis (LDA).}
  \label{tab:accuracyCELEBAvsSINGLEDEEPNETWORK}
\end{table*}

\begin{table}[t]
\centering
\begin{tabular}{c|cccc|}
\cline{2-5}
                                          & \multicolumn{4}{c|}{\textbf{CELEBA Vs DeepNetworks}}                                                                              \\ \cline{2-5} 
\textbf{}                                 & \multicolumn{4}{c|}{\textbf{Kernel Size}}                                                                                        \\
\textbf{}                                 & \textbf{3x3}                       & \textbf{4x4}              & \textbf{5x5}              & \textbf{7x7}                        \\ \hline
\multicolumn{1}{|c|}{\textbf{3-NN}}       & \textbf{89.96}                     & 84.90                     & 80.76                     & 82.69                               \\
\multicolumn{1}{|c|}{\textbf{5-NN}}       & \textbf{90.22}                     & 86.63                     & 82.48                     & 82.77                               \\
\multicolumn{1}{|c|}{\textbf{7-NN}}       & \textbf{89.57}                     & 87.12                     & 82.48                     & 84.27                               \\
\multicolumn{1}{|c|}{\textbf{9-NN}}       & \textbf{89.51}                     & 86.73                     & 83.31                     & 84.27                               \\
\multicolumn{1}{|c|}{\textbf{11-NN}}      & \textbf{89.25}                     & 87.21                     & 83.69                     & 83.97                               \\
\multicolumn{1}{|c|}{\textbf{13-NN}}      & \textbf{89.57}                     & 87.31                     & 84.20                     & 83.45                               \\
\multicolumn{1}{|c|}{\textbf{SVMLinear}}  & 88.02                              & \textbf{88.75}            & 86.05                     & 85.85                               \\
\multicolumn{1}{|c|}{\textbf{SVMsigmoid}} & \multicolumn{1}{l}{\textbf{86.08}} & \multicolumn{1}{l}{72.60} & \multicolumn{1}{l}{83.38} & \multicolumn{1}{l|}{63.66}          \\
\multicolumn{1}{|c|}{\textbf{SVMrbf}}     & \multicolumn{1}{l}{\textbf{89.77}} & \multicolumn{1}{l}{89.71} & \multicolumn{1}{l}{86.24} & \multicolumn{1}{l|}{87.43}          \\
\multicolumn{1}{|c|}{\textbf{SVMPoly}}    & \multicolumn{1}{l}{82.51}          & \multicolumn{1}{l}{86.06} & \multicolumn{1}{l}{84.65} & \multicolumn{1}{l|}{\textbf{86.61}} \\
\multicolumn{1}{|c|}{\textbf{LDA}}        & 87.56                              & \textbf{88.65}            & 86.11                     & 85.48                               \\ \hline
\end{tabular}
\caption{Overall accuracy between CELEBA with all Deep Neural Network, with different kernel size ($3x3$, $4x4$, $5x5$, $7x7$ - obtained through the EM algorithm) and with different classifiers used: KNN, with $k \in \{3,5,7,9,11,13\}$; SVM (linear, sigmoid, rbf, polynomial), Linear Discriminant Analysis (LDA).}
  \label{tab:accuracyCELEBAvsALLDEEPNETWORK}
\end{table}

\begin{table}[t]
\centering
\begin{tabular}{c|cccc|}
\cline{2-5}
                                     & \multicolumn{4}{c|}{\textbf{STYLEGAN Vs STYLEGAN2}}            \\ \cline{2-5} 
\textbf{}                            & \multicolumn{4}{c|}{\textbf{Kernel Size}}                     \\
\textbf{}                            & \textbf{3x3} & \textbf{4x4} & \textbf{5x5}   & \textbf{7x7}   \\ \hline
\multicolumn{1}{|c|}{\textbf{3-NN}}  & 89.36        & 83.57        & \textbf{90.51} & 87.24          \\
\multicolumn{1}{|c|}{\textbf{5-NN}}  & 89.56        & 86.41        & \textbf{89.87} & 85.52          \\
\multicolumn{1}{|c|}{\textbf{7-NN}}  & 89.16        & 85.40        & \textbf{90.93} & 87.59          \\
\multicolumn{1}{|c|}{\textbf{9-NN}}  & 88.55        & 83.98        & \textbf{89.87} & 87.93          \\
\multicolumn{1}{|c|}{\textbf{11-NN}} & 88.35        & 83.37        & \textbf{90.30} & 87.24          \\
\multicolumn{1}{|c|}{\textbf{13-NN}} & 89.36        & 82.76        & \textbf{89.66} & 87.93          \\
\multicolumn{1}{|c|}{\textbf{SVM}}   & 91.77        & 95.13        & 99.16          & \textbf{99.31} \\
\multicolumn{1}{|c|}{\textbf{LDA}}   & 91.16        & 94.52        & \textbf{98.73} & 98.28          \\ \hline
\end{tabular}
\caption{Overall accuracy between STYLEGAN and STYLEGAN2, with all the different kernel size ($3x3$, $4x4$, $5x5$, $7x7$ - obtained through the EM algorithm) and with different employed classifiers: KNN, with $k \in \{3,5,7,9,11,13\}$; Linear SVM, Linear Discriminant Analysis (LDA).}
  \label{tab:accuracyST1ST2}
\end{table}

As described above, the structure of the GAN plays a fundamental role in the Deepfakes detection, in particular for what regards the generator structure. Considering the images from STYLEGAN and STYLEGAN2, it is possible to distinguish them, as the authors of the STYLEGAN2 architecture have only updated parts of the generator in order to remove some imperfections of STYLEGAN. This further confirms the hypothesis,  since even a slight modification of the generator, in particular to the convolution layers, leaves different traces in the images generated. When trying to distinguish the images from STYLEGAN with those of STYLEGAN2, we get a maximum accuracy of the $\textbf{99.31\%}$ (Table~\ref{tab:accuracyST1ST2}).

\begin{figure*}[t]
\begin{center}
\fbox{\rule{0pt}{0in} \rule{.0\linewidth}{0pt}\includegraphics[width=0.70\linewidth]{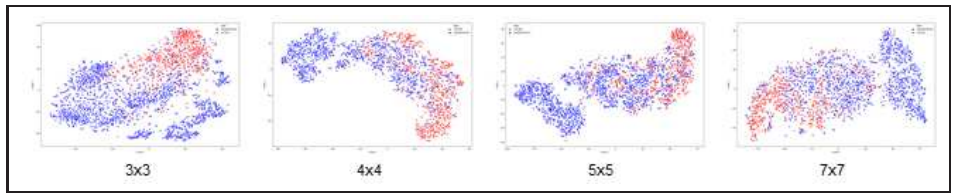}}
\end{center}
   \caption{Two-dimensional t-SNE representation (CELEBA: red; DeepNetwork: blue) of a binary classification problem (with different kernel size): CELEBA Vs DeepNetworks.}
\label{fig:tsneBinary}
\end{figure*}

Finally, another type of classification was the comparison between CELEBA original images and all the images generated with all the networks as a binary classification problem. In this test, a further analysis of the two-dimensional t-SNE was carried out. Figure~\ref{fig:tsneBinary}) shows that, in this case, samples cannot be linearly separated. For this reason, other non-linear classifiers were taken into account reaching a maximum accuracy of $\textbf{90.22}\% $ (with KNN, K=5), with kernel employed in the EM of size $3 \times 3$. Table~\ref{tab:accuracyCELEBAvsALLDEEPNETWORK} shows the obtained results in the binary classification task. 

Many additional experiments were carried out to furtherly demonstrate the effectiveness of the extracted feature vector as a descriptor of the hidden convolutional trace. Specifically results w.r.t. classification tests between different combinations of GANs are described furtherly conferming the robustness of the technique. Also other t-SNE representations are provided and can be found at the following address~\href{https://iplab.dmi.unict.it/mfs/DeepFake/}{https://iplab.dmi.unict.it/mfs/DeepFake/}.

Finally, it is worth to point out that during the research activity a deep neural network technique was employed to detect Deepfakes on the datasets described above. Tests carried out with VGG-16\footnote{\href{https://github.com/1297rohit/VGG16-In-Keras}{https://github.com/1297rohit/VGG16-In-Keras}} on both spatial and frequency domain of images achieved a best result of 53\% of accuracy in the binary classification task showing that a deep learning approach is not able to extract what the proposed approach was able to.
Our results are similar in terms of overall performance by experiments exploited in Wang et al.~\cite{wang2019cnngenerated} that is actually able to reach very high results by simply using a discriminator trained on one family of GANs and using it to infer if images are real or generated from other types of GANs.

\section{Conclusions and future works}
\label{sec:conclusion}
The final result of our study to counter the Deepfake phenomenon was the creation of a new detection method based on features extracted through the EM algorithm. The underlying fingerprint has been proven to be effective to discriminate between images generated by recent GANs architectures specifically devoted to generate realistic people's face. Some more works will be devoted to investigate the role of the kernel dimensions. Also the possibility to extend such methodology to video's analysis and/or evaluate the robustness with respect to standard image editing (e.g. photometric and compression) and malicious processing (e.g. antiforensics) devoted to mask the underlying forensic traces will be considered.
In general one of the key aspect will the possibility to adapt the method in situations on the ``wild" without any a-priori knowledge of the generation process.

\section*{Acknowledgement}
This research was supported by iCTLab s.r.l. - Spin-off of University of Catania (\href{https://www.ictlab.srl}{https://www.ictlab.srl}), which provided domain expertise and computational power that greatly assisted the activity.

\balance
\bibliographystyle{ieee_fullname}
\bibliography{egpaper_final}

\end{document}